\documentclass{article}

\usepackage{microtype}
\usepackage{graphicx}
\usepackage{subfigure}
\usepackage{amsmath}
\usepackage{amssymb}
\usepackage{adjustbox}
\usepackage{multirow}
\usepackage{booktabs} 
\usepackage{caption}

\usepackage{hyperref}

\usepackage[accepted]{icml2019}

\icmltitlerunning{Dissecting Catastrophic Forgetting in Continual Learning by Deep Visualization}

\begin{document}

\twocolumn[
\icmltitle{Dissecting Catastrophic Forgetting in \\
Continual Learning by Deep Visualization}



\icmlsetsymbol{equal}{*}

\begin{icmlauthorlist}
\icmlauthor{Giang Nguyen}{resl}
\icmlauthor{Shuan Chen}{chemical}
\icmlauthor{Thao Do}{resl}
\icmlauthor{Tae Joon Jun}{ansan}
\icmlauthor{Ho-Jin Choi}{resl}
\icmlauthor{Daeyoung Kim}{resl}
\end{icmlauthorlist}

\icmlaffiliation{resl}{School of Computing, KAIST, Daejeon, South Korea}
\icmlaffiliation{ansan}{Ansan Institute for Life Sciences, Ansan Medical Center, Seoul, South Korea}
\icmlaffiliation{chemical}{Department of Chemical and Biomolecular Engineering, KAIST, Daejeon, South Korea}

\icmlcorrespondingauthor{Ho-Jin Choi}{hojinc@kaist.ac.kr}
\icmlcorrespondingauthor{Daeyoung Kim}{kimd@kaist.ac.kr}

\icmlkeywords{Life-long learning, Incremental learning, Explainable AI, Computer vision}

\vskip 0.3in
]




\begin{abstract}
Interpreting the behaviors of Deep Neural Networks (usually considered as a black box) is critical especially when they are now being widely adopted over diverse aspects of human life. Taking the advancements from Explainable Artificial Intelligent, this paper proposes a novel technique called Auto DeepVis to dissect catastrophic forgetting in continual learning. A new method to deal with catastrophic forgetting named critical freezing is also introduced upon investigating the dilemma by Auto DeepVis. Experiments on a captioning model meticulously present how catastrophic forgetting happens, particularly showing which components are forgetting or changing. The effectiveness of our technique is then assessed; and more precisely, critical freezing claims the best performance on both previous and coming tasks over baselines, proving the capability of the investigation. Our techniques could not only be supplementary to existing solutions for completely eradicating catastrophic forgetting for life-long learning but also explainable.
\end{abstract}

\section{Introduction}

Regarding human evolution, life-long learning has been considered as one of the most crucial abilities, helping us develop more complicated skills throughout our lifetime. The idea of this learning strategy is hence deployed extensively in deep learning community. Life-long learning (or continual learning) enables machine learning models to perceive new knowledge while simultaneously exposing backward-forward transfer, non-forgetting, or few-show learning \cite{ling2019unified}. While the aforementioned properties are the ultimate goals for life-long learning systems, catastrophic forgetting or semantic drift naturally occurs in deep neural networks (DNNs) because they are mostly updated upon gradient descent algorithm \cite{goodfellow2013empirical}. 

Many attempts have been succeeded to address the forgetting problem in generative models \cite{zhai2019lifelong}, object detection \cite{shmelkov2017incremental}, semantic segmentation \cite{tasar2019incremental}, or captioning \cite{nguyen2019contcap}. However, algorithms tend to rely on external factors (e.g., input data, network structure) while ignoring why catastrophic forgetting happens internally. If we have the picture of the forgetting process and understand how this problem affects models, it would be one step towards learning without forgetting. 

Contemporary interpretability methods give us advantages to understand the decision-making process of deep neural networks, ranging from visualizing saliency maps \cite{simonyan2013deep, dabkowski2017real} to transforming models into human-friendly structures \cite{che2016interpretable}. Interpreting the activation of a neuron or a layer in networks helps us categorize the specific role of each block, layer, or even a node. It has been proven that the earlier layers extract the basic features, such as edges or colors; while deeper layers are responsible for detecting distinctive characteristics. Prediction Difference Analysis (PDA) \cite{zintgraf2017visualizing}, even more specifically, highlights pixels that support or counteract a certain class, indicating which features are positive or negative to a prediction.  

Although catastrophic forgetting is tough and undesirable, research to understand this problem is rare amongst AI community. The interest in understanding or measuring catastrophic forgetting does not correlate with the number of research to deal with this problem. \cite{kemker2018measuring} develop new metrics to help compare continual learning techniques fairly and directly. \cite{nguyen2019toward} study which properties cause the hardness for the learning process. By modeling the chosen properties using task space, they can estimate how much a model forgets in a sequential learning scenario, shedding light on factors affecting the error rate on a task sequence. However, they can not explain what is being forgotten or which components are forgetting inside the model, but showing what properties of tasks trigger catastrophic forgetting. In comparison, our work focuses on study which components of a network are most likely to change corresponding to a given sequence of tasks.

This research introduces a novel approach to dissect catastrophic forgetting by visualizing hidden layers in class-incremental learning (considered as the hardest learning scenario in continual learning). In this learning paradigm, rehearsal strategies using previous data are prohibited and samples of the incoming tasks are unseen so far. We propose a tool named \texttt{Auto DeepVis} which leverages the prediction difference analysis from \cite{zintgraf2017visualizing}. This tool automates the dissection of catastrophic forgetting, exactly pointing out which components in a model are causing the forgetting. By using Intersection over Union (IoU), the degree of forgetting is measured after each class is added, thus giving us an intuition how forgetting happens on a given part of the network.

In the first step, our tool observes the evidence and against for a prediction. If the evidence for a given image changes, we argue that the model is forgetting what it needs to look. IoUs between the previous and current evidence are taken into account to determine the degree of forgetting. We automate this procedure on representative samples of trained classes, followed by a generalization step to figure out the main culprits of forgetting. The algorithm is outlined in Algorithm \ref{alg:autodeepvis}.

In our thorough analysis of the model, it is necessary to pick the appropriate components in a block to visualize the changes. For instance, in ResNet-50 \cite{he2016deep}, there are 5 convolutional blocks, thus interpreting the activation of all filters in the whole 5 blocks will take a large amount of time. We propose to choose the filter having the highest IoU value with the ground truth segmentation. This choice is sufficient because although some filters may be blocked by activation function, ranking the importance of the remaining filters over a convolutional block is impossible. As a result, the biggest-IoU filter can be the representative for a given block.

A work from \cite{kemker2018measuring} conducts experiments on state-of-the-art continual learning techniques that address catastrophic forgetting. It is demonstrated that although the algorithms work, but only on weak constrains and unfair baselines, thus the forgetting problem is not yet fully solved. They insist on the infeasibility of using toy datasets, such as MNIST \cite{kirkpatrick2017overcoming} or CIFAR \cite{zenke2017continual} in continual learning. Consequently, we choose Split MS-COCO \cite{nguyen2019contcap} to measure the forgetting on deep neural networks. From the results of the dissection, we simply freeze the most plastic components in the network to protect the accumulated information. In addition to visualization on Convolutional Neural Network (CNN), we also briefly study how the decoder which consists of Recurrent Neural Network (RNN) changes in continual learning.

We summarize the contributions of this work as follows: First, we propose a novel and pioneering method to analyze catastrophic forgetting in continual learning. \texttt{Auto DeepVis} automatically points out the forgetting components in a network while learning a task sequence. Second, we introduce a new approach to mitigate catastrophic forgetting based on findings. Our techniques could play a complementary role in the step towards eradicating catastrophic forgetting.

\section{Related Work}

\paragraph{Feature Visualization}
To understand how a model forgets the features learned before, we need to visualize how the model processes the input image. To do this, the layer-wise visualization of the model is performed. The old model (or original model) is the model we have obtained so far, denoting acquired knowledge on past tasks. On the other hand, the new model (or current model) is the network facing the new incoming tasks and should avoid catastrophic forgetting.

There are several works trying to visualize convolutional neural networks by various kinds of means. \cite{zeiler2014visualizing} make use of multi-layered deconvolutional networks (deconvnet), using switches to record the local max value of the original input images. Deconvnet allows us to recognize which features are expected by a specific part of a network or what properties of image excite a chosen neuron the most. Instead of feeding an input image to diagnose, \cite{yosinski2015understanding} attempt to generate a synthetic image which maximizes the activation of a given neuron by gradient descent algorithm. 

Another broadly used approach called saliency maps by \cite{simonyan2013deep}, presenting a gradient-based technique to measure how the classification score is sensitively changed due to the small changes from pixels. This work shows a generalized version of deconvolution layers. A similar approach has been proposed by \cite{robnik2008explaining}, which completely removes input pixels instead of calculating the gradient of those pixels. A variant of the above-mentioned work called PDA is proposed by \cite{zintgraf2017visualizing}. They utilize conditional sampling that sweeps away patches of connected pixels instead of removing one single pixel at a time, which eventually shows better visualization results. However, this tool only generates the computer vision, leaving the conclusion for users. This manual process can not ensure the quality of the observation when we can have hundreds or even thousands of feature maps in a convolutional block. In this work, PDA is adopted to build an automatic tool for visualizing catastrophic forgetting.

\paragraph{Catastrophic Forgetting}
Catastrophic forgetting has been introduced for the first time by \cite{mccloskey1989catastrophic} in connectionist networks. Very recently, \cite{toneva2018empirical} reframe the definition of forgetting as when the prediction of a model on a sample is shifted during the learning process, from correctly to incorrectly. Going deeply into the model, we can argue that the values of neurons have changed drastically, thus possibly resulting in a different answer for the same input image.  

Elastic Weight Consolidation - EWC \cite{kirkpatrick2017overcoming} uses Fisher information matrix to help model mimic the synaptic consolidation mechanism of the human brain. Important parameters for the performance of the old task are protected while others are updated to minimize the loss on the new dataset. By comparison, the significance of each synapse (or weight) in the neural network is computed locally in \cite{zenke2017continual}. When the distribution interference appears, synaptic states keep and estimate the importance of synapses by an online estimation and crucial synapses are prevented from changing. Learning without Forgetting - LwF \cite{li2017learning} generates pseudo labels on incoming data to help capture the previous distribution. In training, knowledge distillation \cite{hinton2015distilling} turns the pseudo labels into soft targets and a warm-up step is applied. Based on Bayesian neural networks, \cite{lee2017overcoming} match the moment of posteriors on both two tasks to guide the new network to a common low-error region. Knowledge distillation from the old model and an expert feature extractor is conducted in \cite{hou2018lifelong}, complemented by a retrospection on a trivial fraction of old data. \cite{rusu2016progressive} dynamically expand the network by specialized sub-networks to absorb new knowledge while the old modules are frozen.

Catastrophic forgetting can be generally addressed by regularization, data replay or altering network architecture. In this research, we show that only by simply freezing the fragile blocks of a network, we can significantly improve the generalization performance.

\begin{figure}[h!]
  \includegraphics[scale=0.85]{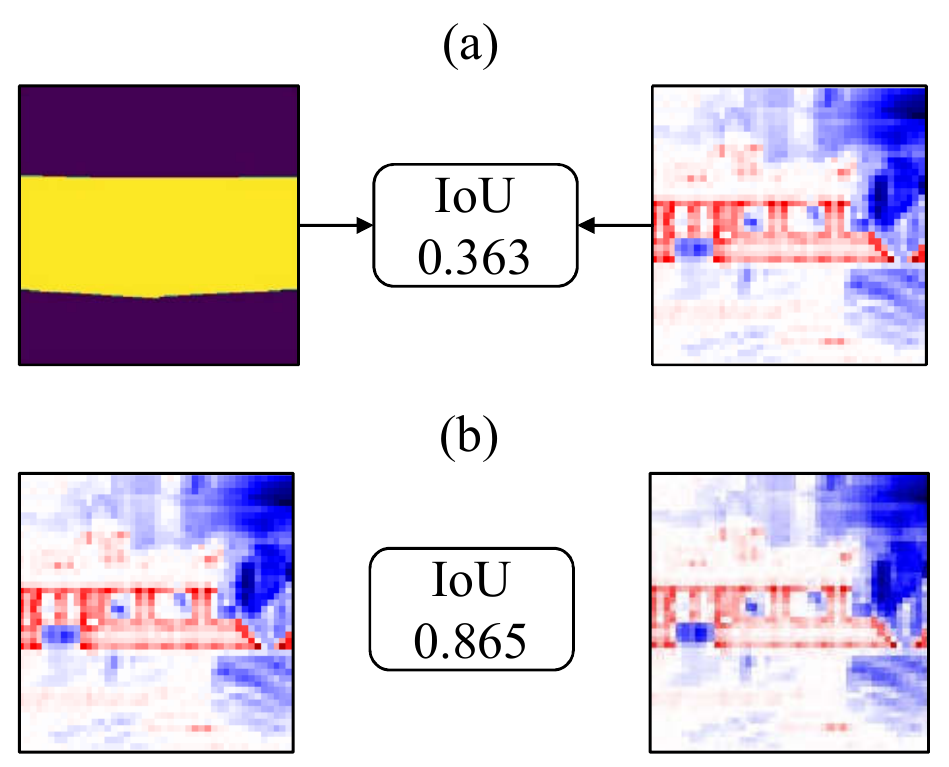}
  \caption{(a) IoU value between the segmentation of a train and the positive features. (b) The IoU of two representative maps. Red is evidence and blue is against.}
  \label{fig:iou1}
\end{figure}

\section{Methodology}
Although \cite{nguyen2019toward} show an interest in understanding catastrophic forgetting, they focus on how task properties influence the hardness of sequential learning. Hence, they are explaining based on the input data. In comparison, our work attempts to explain how the forgetting happens over time based on the computer vision of models. \cite{zintgraf2017visualizing} present response from a network to a given image in which we can clarify which features support or counteract the prediction. We leverage this tool for visualizing the computer vision, but extend to an automatic version - \texttt{Auto DeepVis} to efficiently dissect the forgetting dilemma. The ultimate goal of this tool is to figure out the most plastic layers or blocks in a network. Plasticity means low degree of stiffness or easy to change. Moreover, continual learning techniques have been proposed to alleviate catastrophic forgetting, but none of them are devised based on the findings from Explainable AI. Critical freezing is built on the top of \texttt{Auto DeepVis}' investigation to provide an interpretable yet effective approach to acquire the life-long learning ability for deep learning models. In the learning process, the stable state of the old model is employed to initialize the new network. This way really mimics the working mechanism of the human brain.

\begin{figure*}[h!]
  \centering
  \includegraphics[scale=0.5]{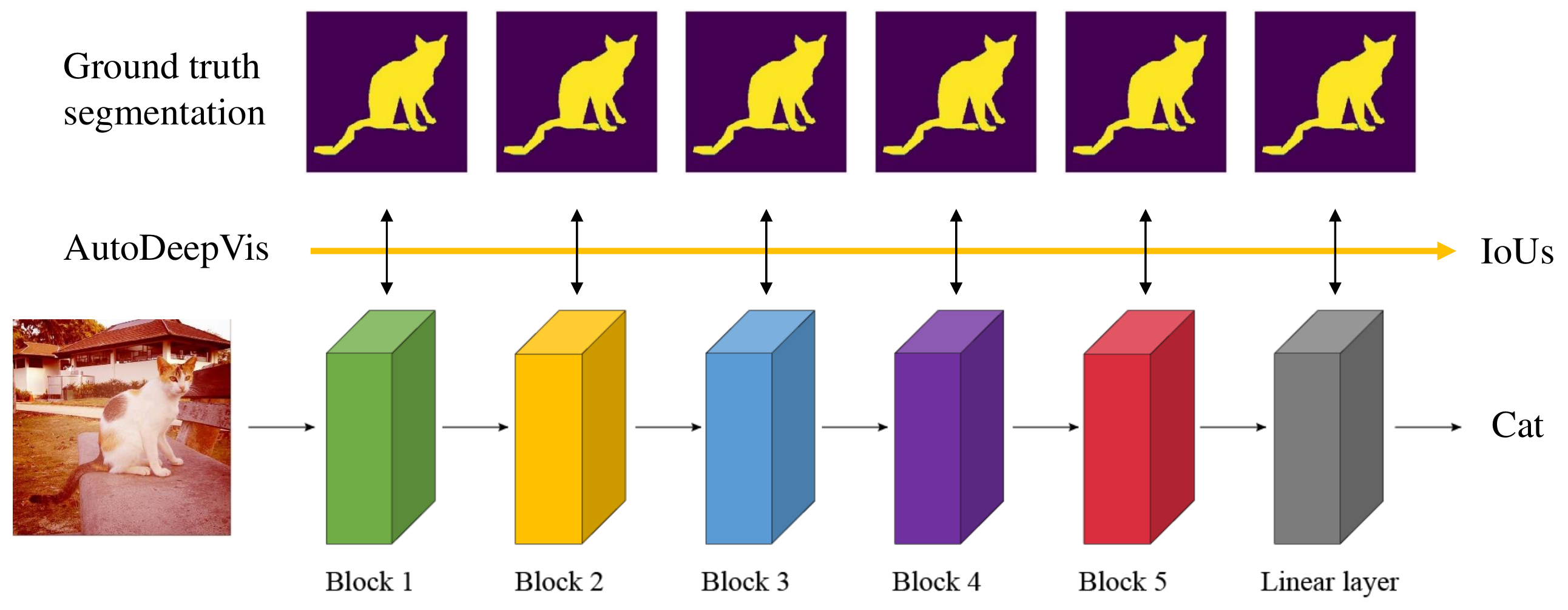}
  \caption{Overall scheme of Auto DeepVis}
  \label{fig:AutoDeepVis}
\end{figure*}

\subsection{Auto DeepVis}
To dissect the model, we visualize the hidden layers to understand the forgetting effect inside a model after being trained on different tasks. \cite{zintgraf2017visualizing} claim that different features of the objects being dissected could be captured and visualized by particular feature maps in different layers. By looking into every response maps in one convolutional block, we realize diverse features, such as eyes, face shape, car wheels, or background are isolatedly recognized by different channels. Unfortunately, finding each feature map manually by human eyes might be inefficient. To solve this issue, we only compare the computer vision with the ground truth segmentation rather than small details. In general, we do not look for the answer that what features are being forgotten, but which layers are now forgetting.

Assume the semantic segmentation label of MS-COCO dataset \cite{lin2014microsoft} is what human sees, we compare this segmentation with the computer vision of the model, particularly concentrating on positive evidence for a prediction. The IoU value between the segmentation and evidence is calculated as shown in Fig. \ref{fig:iou1} (a). The red dots in the map describe the positive evidence while the blue points represent the against. The scheme of \texttt{Auto DeepVis} is illustrated in Fig. \ref{fig:AutoDeepVis}.

Having the \textit{m-th} feature map (FM) in the \textit{l-th} layer of a model \textit{M} and the ground truth segmentation \textit{GT}, the IoU is computed as:
\begin{equation}
\label{eq:1}
IoU_{M,GT} (l,m) =  \frac{FM(m,l) \cap GT }{FM(m,l) \cup GT}
\end{equation}

To select the feature map having the largest overlap with the ground truth in each convolutional block, the representative feature map (RM) with the best IoU is denoted as $RM_{M,GT}$ :
\begin{equation}
\label{eq:2}
RM_{M,GT}  (l) = argmax_m (IoU_{M,GT} (l,m))
\end{equation}

To understand how the computer vision change over the training process, we compared the RM in the new model with the RM of original network $RM_{M_O,GT}$ Fig. \ref{fig:iou1} (b). The forgetting effect of each trained model is measured by the IoU between the original model \textit{$M_O$} and new model \textit{$M_N$}:
\begin{equation}
\label{eq:3}
IoU_{M_N,M_O} (l,m) = \frac{F M(m) \cap RM_{M_O,GT} }{F M(m) \cup RM_{M_O,GT}}
\end{equation}

Similar to the method of finding out the best map fitting with ground truth, the feature map representing the best memory of the original feature map is denoted as $RM_{M_N,M_O}$ in Equation \ref{eq:4}. In the same block of both the old and new model, the role of a filter can be adjusted. For instance, the $50^{th}$ filter in the $2^{nd}$ block of the old model detects the eyes, but the same filter in the same block of the new model will consider the face. The way we propose to pick $RM_{M_N,M_O}$ sounds heuristically sufficient.
    
\begin{equation}
\label{eq:4}
RM_{M_N,M_O}  (l) = argmax_m (IoU_{M_N,M_O} (l,m))
\end{equation}

\begin{algorithm}[tb]
   \caption{Auto DeepVis}
   \label{alg:autodeepvis}
\begin{algorithmic}
   \STATE {\bfseries Input:} Sample set $S$, segmentation ground truth $GT$, old model $M_O$, new model $M_N$, number of blocks $K$
   \STATE {\bfseries Output:} Forgetting layers $\mathbb{F}$
   \STATE $i=0$
   \STATE $\L = \O$
   \REPEAT
   \STATE $I = S[i]$
   \STATE $IoUs = \O$
   \STATE $RM = \O$
   \STATE $FM = PDA(I)$
   \FOR{$j=1$ {\bfseries to} $K$}
   \STATE $RM_{M_O,GT} \leftarrow \textit{FM with highest } {IoU_{M_O,GT}}$
   \STATE $RM_{M_O,M_N} \leftarrow \textit{FM with highest } {IoU_{M_O,M_N}}$
   \STATE \textbf{Append}($IoUs, max(IoU_{M_O,M_N})$)
   \STATE \textbf{Append}($RM, RM_{M_O,M_N}$)
   \ENDFOR
   \STATE $\boldsymbol{b} \leftarrow \textit{block with highest drop in IoUs}$ 
   \STATE \textbf{Append}(\L, $\boldsymbol{b}$)
   \STATE $i = i+1$
   \UNTIL{$i =  \textbf{size}($S$)$}
   \STATE $\mathbb{F}$ $\leftarrow$ \textit{Most frequent block in} \L
\end{algorithmic}
\end{algorithm}

The pipeline of our method is depicted in Algorithm \ref{alg:autodeepvis}. The sample set $S$ is particularized in Section \ref{sec:exp}, and K is the number of the convolutional blocks in the network. By inputting image by image from $S$, we get the visualization of filters over the whole network by PDA. Next, we iterate through the blocks to obtain the fragile block against a given image, then the \textbf{indexOf} gets the most forgetting block. Finally, we generalize on $S$ to return the most forgetting component $\mathbb{F}$. 

 \begin{figure*}[h!]
	\centering
	\includegraphics[scale=0.6]{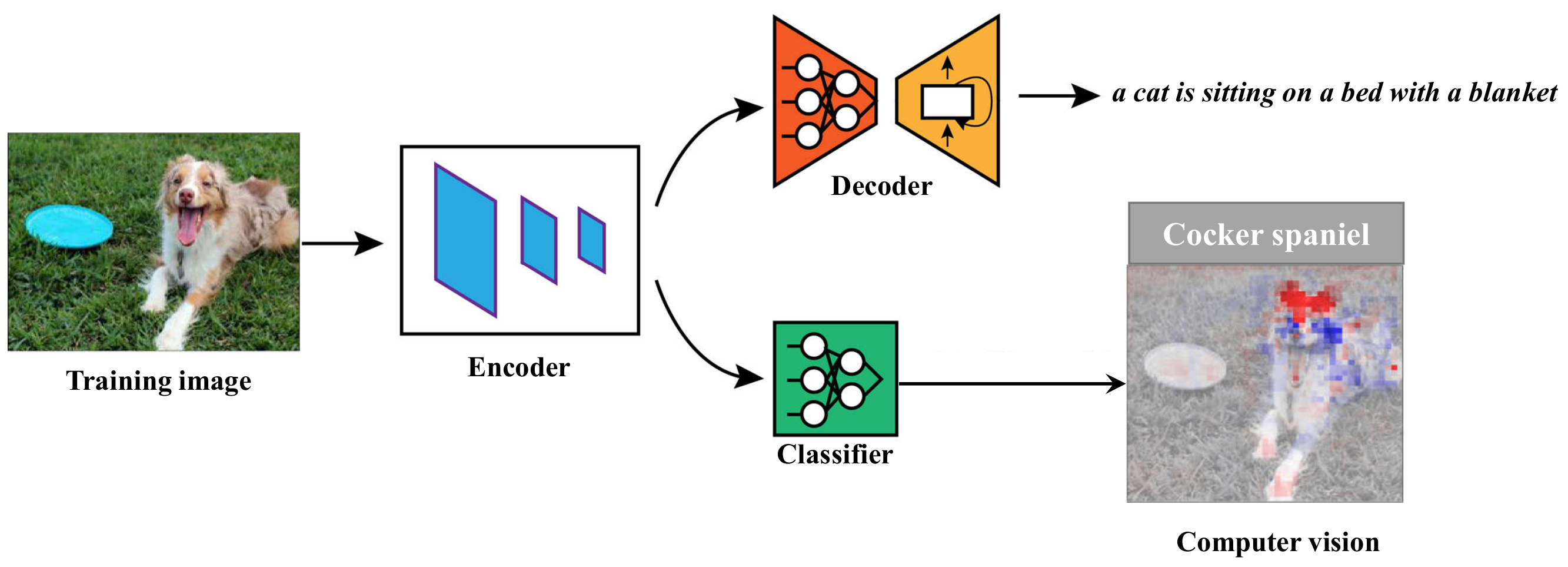}
	\caption{Two-head network for dissecting catastrophic forgetting.}
	\label{fig:two-head}
\end{figure*}

\subsection{Critical Freezing}
Using the investigation from \texttt{Auto DeepVis}, we freeze or apply a tiny learning rate on the most plastic layer in a deep learning model. Take Resnet-50 including 5 convolutional blocks as an example, if we find the $3^{rd}$ convolutional layer fragile, this layers should be slightly updated or completely frozen in the training process of the next task. The objective function is the standard cross-entropy loss for image captioning, in which $\mathbb{V}$ is the size of the vocabulary.

\begin{equation}
\begin{aligned}
	\mathcal{L} = L_{CE}
	            = - \sum_{i=1}^{\mathbb{V}} Y_{k}^{i} \log \hat{Y}_{k}^i
\end{aligned}
\end{equation}

The proposed technique can accompany various existing solutions. For instance, when using knowledge distillation for continual semantic segmentation in \cite{michieli2019incremental}, progressively adding classes alters the evolution of the model, misleading the network to a local optimum. By knowing which regions are needed to be intact, the performance on the old task could be largely improved. Our tool could also benefits other continual learning approaches, ranging from regularization to dynamic architecture.

\begin{table*}[htb!]
\caption{Performance when 5 classes arrive sequentially on past tasks and newly added tasks.}
\begin{center}
\begin{adjustbox}{width=1.0\textwidth,center=\textwidth}
\begin{tabular}{|l|c|c|c|c|c|c|c|c|c|c|}
\hline
\multirow{2}{*}{} & \multicolumn{5}{c|}{$Past-tasks$}                         & 
\multicolumn{5}{c|}{$New-tasks$}              \\ \cline{2-11} 
                  & BLEU1          & BLEU4          & METEOR         & ROUGE\_L       & CIDEr          & BLEU1          & BLEU4          & METEOR         & ROUGE\_L       & CIDEr          \\ \hline
$Original$               & 68.1         & 24.9        & 23.4         & 50.8         & 77.8         & -              & -              & -              & -              & -              \\ \hline
$Fine-tuning$       & 46.0          & 6.5          & 11.7          & 34.5          & 11.3          & 58.1          & 15.5          & 17.7          & 44.2         & 35.0          \\ \hline
$Encoder-Freezing$    & 51.3         & 11.1          & 15.3        & 38.8          & 27.1          & 60.5          & 17.2          & 19.3          & 45.4          & 43.7          \\ \hline
$Decoder-Freezing$    & 53.3 & \textbf{13.6} & 15.6 & 40.3 & \textbf{35.3} & 60.2 & 17.3 & 17.8 & 44.7 & 36.5 \\ \hline
$Critical-Freezing$     & \textbf{54.3}          & 12.5         & \textbf{16.1}         & \textbf{40.4}         & 33.0          & \textbf{61.1}          & \textbf{17.8}          & \textbf{19.6}          & \textbf{46.3}          & \textbf{45.6}          \\ \hline
\end{tabular}
\end{adjustbox}

\label{tab:my-table3}
\end{center}
\end{table*}

\section{Experiments}
\label{sec:exp}
 When using PDA \cite{zintgraf2017visualizing}, the optimal window size for the best visualization is $k=5$. We use a tweaked dataset called Split MS-COCO from \cite{nguyen2019contcap} to reproduce catastrophic forgetting. The dataset contains over 47k images for training and over 23k images for validation and testing. For incremental learning setup, at one time step, a new class will arrive. Notably, data balancing is not applied yet; as a result, this technique could be left for future work to increase the overall performance. 
 
 Experiments are performed on a multi-modal task (captioning) combining both CNN and RNN in the architecture, obeying the sequential scenario from \cite{nguyen2019contcap}. We initially train 19 classes to acquire the original model, then adding 5 classes incrementally. Obviously, if the investigation from our tool makes the chosen task work, the findings could be also deployed on object detection, classification or segmentation. The captioning model is divided into an encoder followed by a decoder. Therefore, the original structure should be transformed into a two-head network so that we can get the prediction of the CNN and the sentence generated from the decoder simultaneously. We add one more output layer as a classifier after the encoder, and our proposed architecture is presented in Fig. \ref{fig:two-head}. The encoder is the ResNet-50, and the decoder includes an embedding layer, a single-layer LSTM, and a fully-connected layer producing a word at a time step.
 
 \begin{figure}[hb!]
	\centering
	\includegraphics[width=0.5\textwidth]{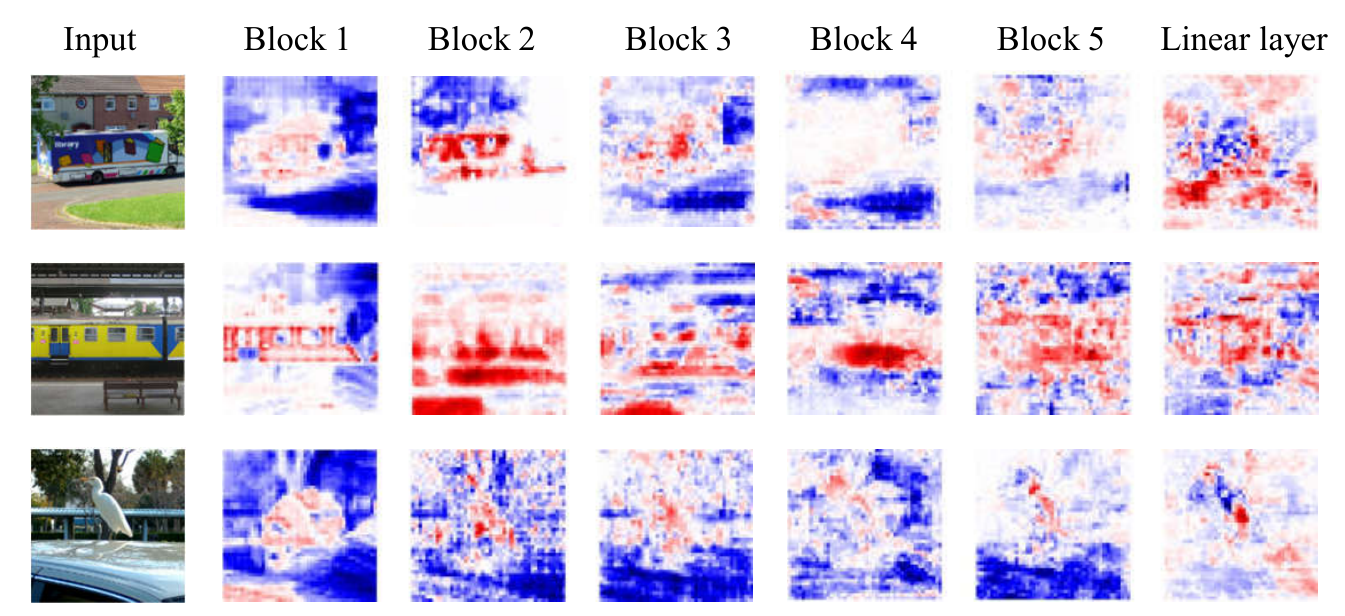}
	\caption{Feature maps from convolutional blocks of ResNet-50.}
	\label{fig:Visualizing}
\end{figure}
 
 As our tool works on a single image, running multiple times on different and diverse input images is needed, helping us generalize the forgetting. We choose a sample set $S$ (\textit{bicycle, car, motorcycle, airplane, bus, train, bird, cat, dog, horse, sheep, and cow}) from 19 trained classes.
 
To evaluate critical freezing, there are baselines, such as fine-tuning or heuristically freezing. In fine-tuning, The old model initializes the new model and training is done to minimize the loss on the new task. As the network contains two parts, encoder and decoder, we freeze them separately. The traditional scores for image captioning are considered in evaluating the superiority of critical freezing over the baselines.  

\subsection{Auto DeepVis to Dissect CNNs}
To dissect which parts in the encoder are forgetting the most, we first apply \texttt{Auto DeepVis} to elaborate the IoU of each layer comparing with the ground truth and the original computer vision seen by layers of the original model $M_{19}$ (trained on 19 classes). After adding a new class, we obtain $M_{20}$, and $M_{n}$ is the model when $n$ classes are witnessed. In Fig. \ref{fig:Visualizing}, the visualized results show that the first and second block of ResNet-50 can overall capture the outline of objects. Computer vision turns to represent more detailed features from the objects and other background features to determine the class of the input image.

\begin{figure}[hbt!]
	\centering
	\includegraphics[width=0.48\textwidth]{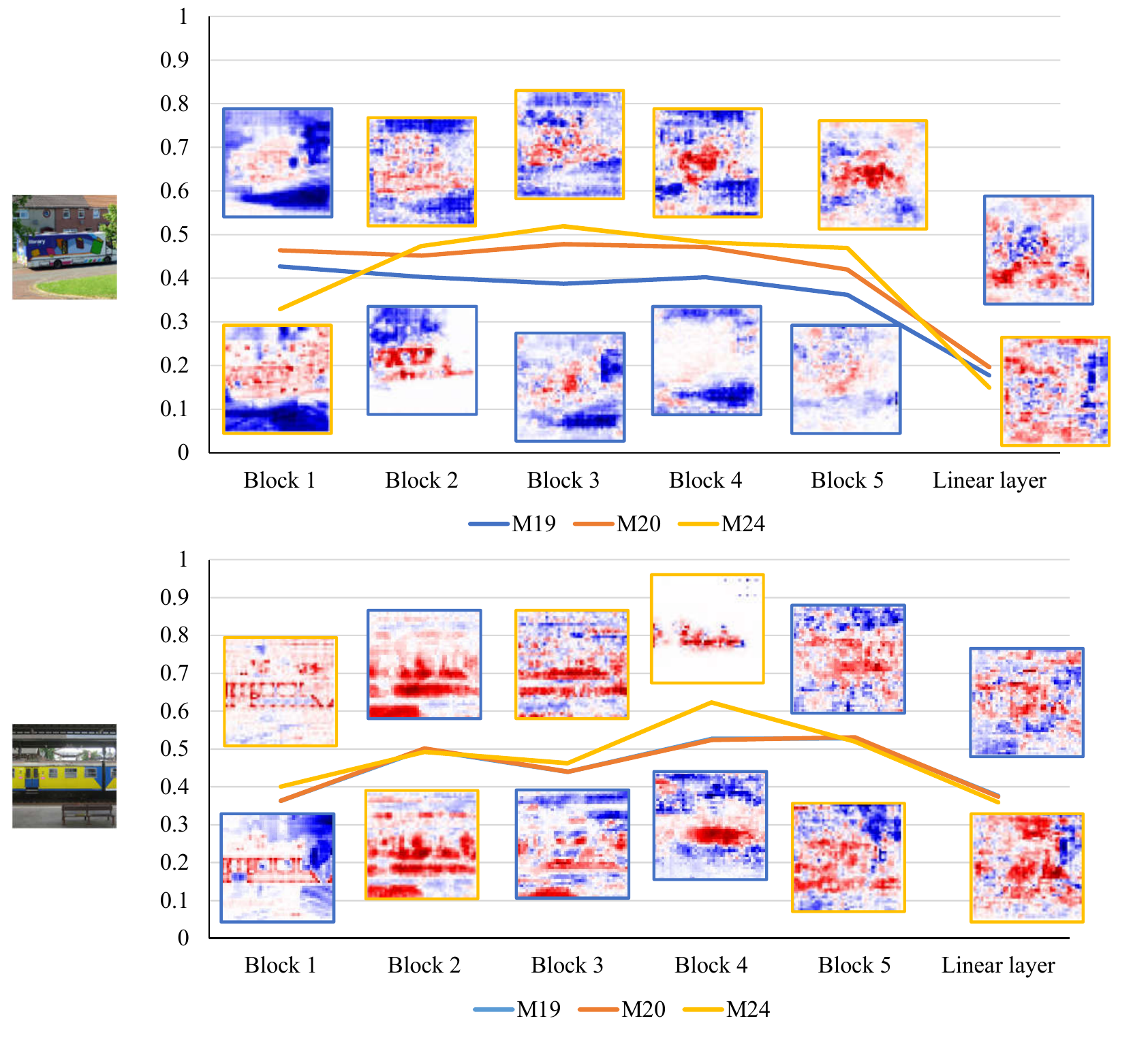}
	\caption{$IoU_{M,GT}$ of $M_{19}$, $M_{20}$ and $M_{24}$ comparing with GT.}
	\label{fig:IoU_GT}
\end{figure}

Subsequently, the IoUs of different blocks in each model, comparing with ground truth, are calculated by equation \ref{eq:1} and \ref{eq:2}. The results reveal that although different models show different levels of performance of classification, the $IoU_{M_n,GT}$ ($n>19$) values are roughly similar at all the layers, which implies that no matter the how good the performance is, the level of matching between feature maps of each model and the human vision is preserved (Fig. \ref{fig:IoU_GT}).

\begin{figure}[h!]
	\centering
	\includegraphics[width=0.48\textwidth]{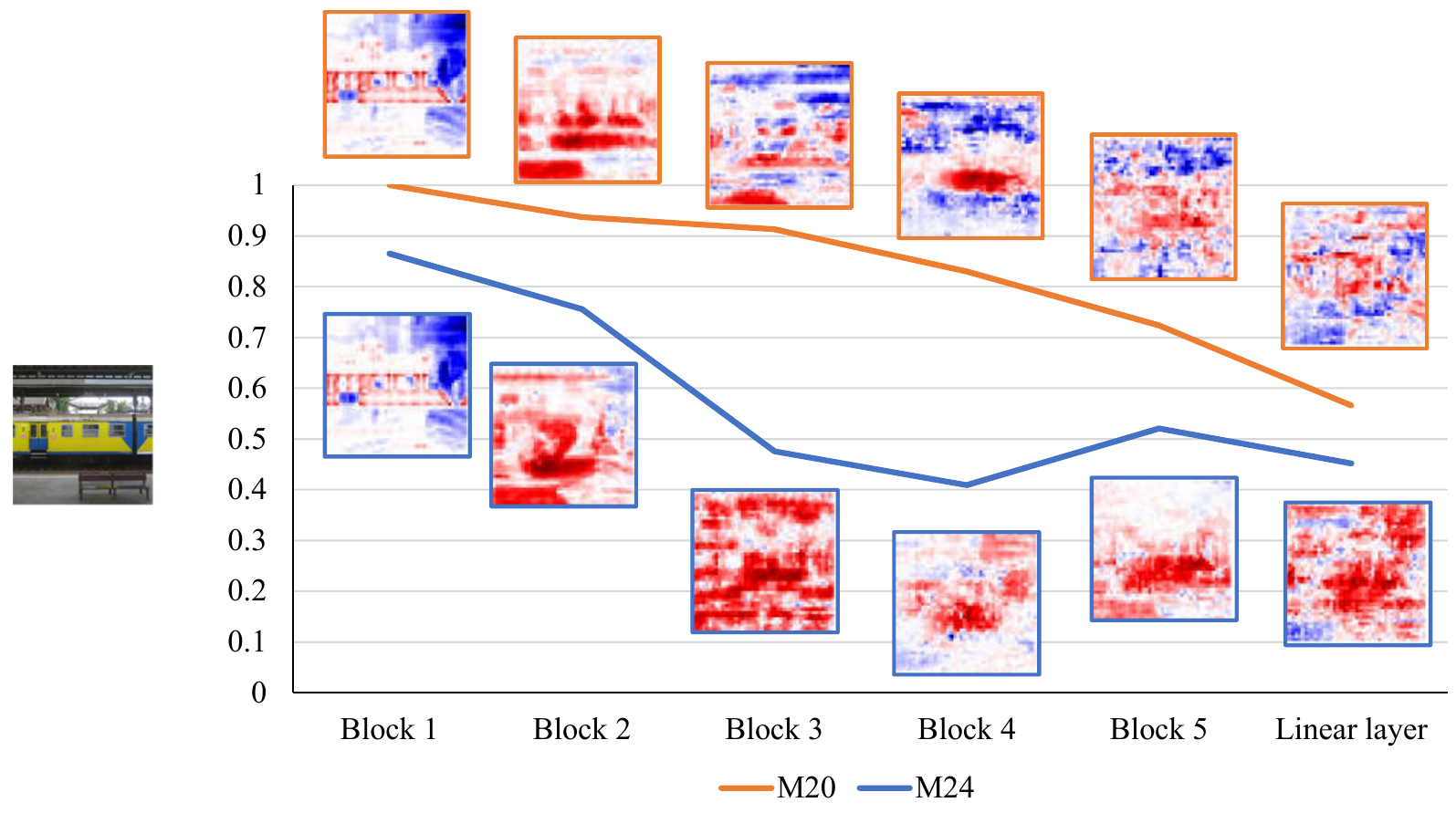}
	\caption{$IoU_{MT,MO}$ of model $M_{20}$ and $M_{24}$ comparing with $M_{19}$.}
	\label{fig:IoU_M19}
\end{figure}

To quantitatively measure how the forgetting occurs in the encoder, we compute IoUs by equation \ref{eq:3} and \ref{eq:4}. As shown in Fig \ref{fig:IoU_M19}, it is clear that the IoU between $M_{20}$ and $M_{19}$ is much higher than the figure for $M_{24}$ and $M_{19}$ in every block. For $M_{20}$, the $IoU_{M_{20},M_{19}}$ is always 1.00 at the first block of the network, showing that a very trivial forgetting happens here. The $IoU_{M_{20},M_{19}}$ starts to drop along the blocks because the later feature maps are constructed by the previous maps. The forgetting effect persists and does not show which block is forgetting the most. 

\begin{table}[htb]
	\caption{Qualitative analysis on the encoder. The prediction is preserved via critical freezing.}
    \label{tab:quali} 
	\includegraphics[width=0.48\textwidth]{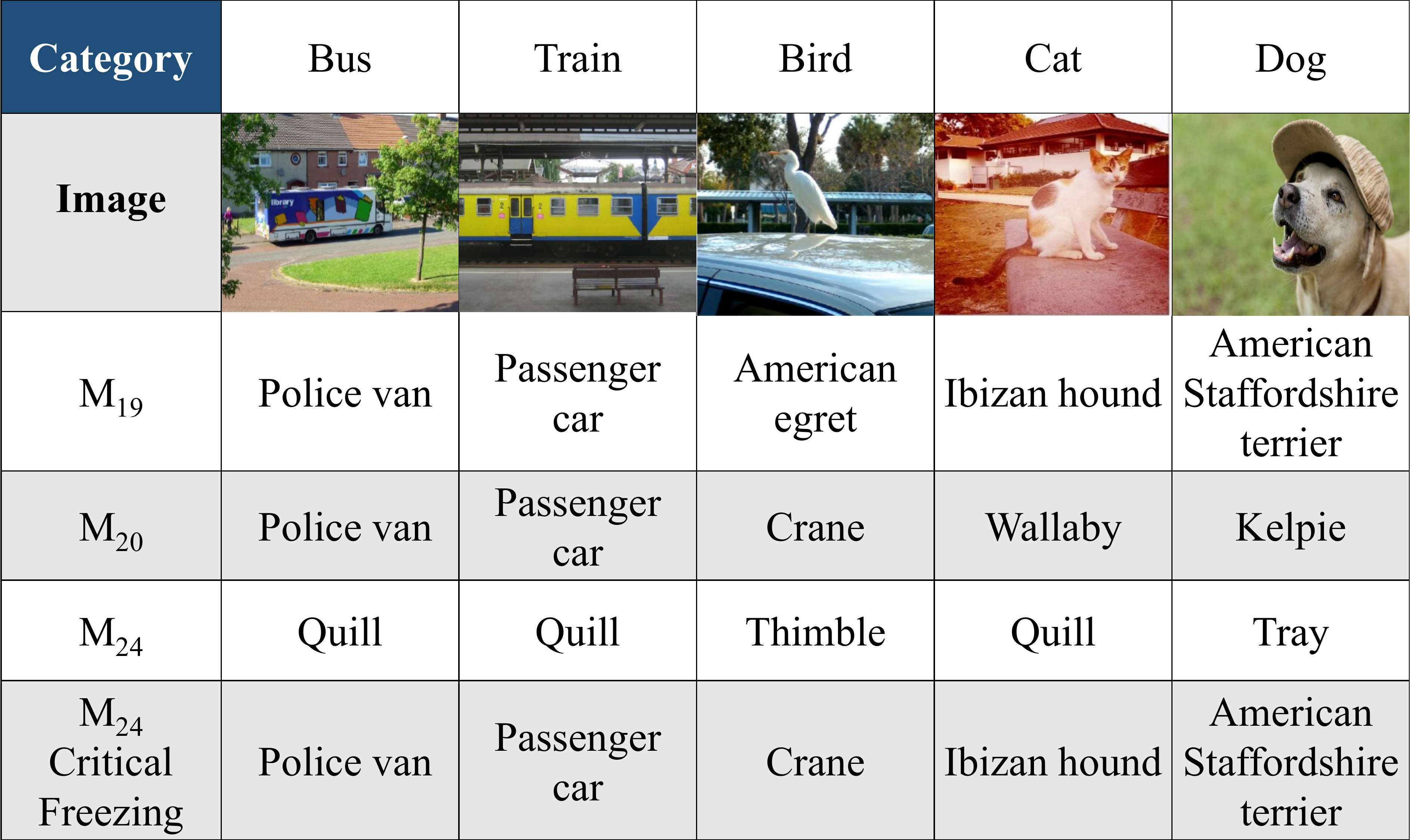}
\end{table}

For $M_{24}$, the first block of the model still gets a high IoU comparing with $M_{19}$ and the values decrease from the second block.
Unlike the stably decreasing trend seen in $M_{20}$, the decreasing rate of $IoU_{M_{24}, M_{19}}$ fluctuates through the blocks and a severe drop at block 3 is observed in all the testing input, suggesting that the forgetting effect might happen the most in this block. After the $3^rd$ block, almost instances show a continuous IoU decline at block 4. Iterating this procedure on all images of the set $S$ reinforces that the most dramatically forgetting is happening at block 3 and block 4 in ResNet-50.

Freezing the mentioned blocks means the feature extraction stays unchanged. We conduct a qualitative analysis to observe the response of the encoder while critical freezing is adopted in Table. \ref{tab:quali}. The prediction keeps virtually the same with the output of the original model. The only misclassifcation in the Table. \ref{tab:quali} is from \textit{American eagle} and \textit{Crane}; however, they are still from the bird family. While two naive approaches of freezing in \cite{nguyen2019contcap} are also implemented (partially freezing), we devise critical freezing based on findings, which only freezes the most plastic layers. As shown in Table. \ref{tab:my-table3}, precisely freezing helps learning on both the new and old tasks more effectively.

\subsection{Decoder Dissection}
In addition to dissecting CNN, we also want to inspect and visualize the changes of the decoder. However, to the best of our knowledge, there is still no work done in revealing how RNNs visually see or sense the input. Therefore, in this work, we simply freeze each component of the decoder in learning the new task and observe the effect of freezing. We also keep the encoder frozen from $M_{19}$ to see how the synthetic caption depends on an individual layer of the decoder. Table \ref{tab:my-table4} suggests the significance of the LSTM network in the sentence generation process while the linear layer can be trainable over a task series. 

\begin{table}[]
\caption{Performance of $M_{24}$ on the past tasks when freezing decoder's components.}
\resizebox{\columnwidth}{!}{%
\begin{tabular}{|c|c|c|c|c|c|}
          & BLEU1 & BLEU4 & METEOR & ROUGE\_L & CIDEr \\
LSTM      & 46.1  & 7.0   & 11.1   & 34.1     & 12.8  \\
Embedding & 45.0  & 6.7   & 11.1   & 34.0     & 12.8  \\
Linear    & 39.3  & 2.1   & 8.3    & 31.3     & 4.2  
\end{tabular}
}
\label{tab:my-table4}
\end{table}

\section{Conclusion}
As the presence of catastrophic forgetting hinders the life-long learning, understanding how this phenomenon happens in computer vision is extremely significant. We introduce \texttt{Auto DeepVis} to grasp catastrophic forgetting. From knowing where the forgetting issue is coming from, a technique has been proposed focusing on plastic components of a model to moderate the information loss. The results indicate the superiority of critical freezing over the baselines. We also try knowledge distillation on plastic layers but it does not help much because the discrepancy of teacher and student is accumulated over time, leading to a difficulty for teaching. To the best of our knowledge, no work has been done for mitigating catastrophic forgetting based on Explainable AI. This work shows a satisfying result from the investigation. \texttt{Auto DeepVis} gives good observation on the forgetting layer, and freezing critical layers helps the model mitigate catastrophic forgetting and could play as a supplement to other techniques to completely address catastrophic forgetting. 

There are future works following our paper. First and foremost, a deeper, cleaner, and effort-free version of our tool should be taken into consideration to give a better insight into catastrophic forgetting. At this time, we only consider the filter having the highest IoU. Although this assumption is appropriate, it is still crude, thus looking into other filters in one convolutional block might be necessary as well. Scaling this work for other tasks can better validate the feasibility of the proposed continual learning algorithm. Last but not least, RNN is now being overlooked and not understood fully via interpretability methods. RNN dissection requires efforts but would open the door to understanding the nature of this network.

\bibliography{example_paper}
\bibliographystyle{icml2019}

\end{document}